\documentclass[letterpaper, 10 pt, conference]{ieeeconf}  
\usepackage[T1]{fontenc}
\usepackage{amsmath}
\usepackage{amssymb}
\usepackage{algorithm}
\usepackage{algpseudocode}

\let\labelindent\relax
\usepackage{enumitem}
\usepackage{cuted}

\usepackage{graphicx}
\usepackage{array}
\usepackage{tabulary}
\usepackage{multirow}
\usepackage[table]{xcolor}
\usepackage{ctable}
\usepackage{booktabs}		
\usepackage{threeparttable}	
\usepackage{cite}
\usepackage{hyperref}

\IEEEoverridecommandlockouts                              

\makeatletter
\def\fnum@figure{Figure~\thefigure} 
\makeatother

\title{\LARGE \bf
Contact-Aware Modality Augmentation \\ for Fine-Tuning Robot Foundation Policies
}

\author{
Junsung Park$^{1}$, Hogun Kee$^{1}$, Songhwai Oh$^{1}$%
\\[2mm]
$^{1}$Seoul National University
}

\begin{document}
\bstctlcite{IEEEexample:BSTcontrol}

\maketitle
\thispagestyle{empty}
\pagestyle{empty}

\begin{abstract}

Robot foundation policies are pretrained on large-scale demonstrations, but most of this data contains only RGB images and proprioception. These policies therefore rely on images to detect when the hand grasps an object or when the object slips, which is difficult when the hand occludes the object. Contact signals can complement images in such cases, but it remains an open question where they should enter a pretrained policy so that it can use them effectively without losing what it has already learned. In this work, we study how a robot foundation policy pretrained without contact can learn to use it during fine-tuning. We compare state fusion, which concatenates the contact signal to the proprioceptive state, with a dedicated contact encoder, which gives the encoded contact signal to the action head as a new token. We fine-tune GR00T N1.5 using each method and evaluate the resulting policies in simulation on two humanoid robots, GR1 and Unitree G1. Since existing humanoid datasets rarely record contact, we augment a public GR1 dataset with binary contact signals and build a simulation pipeline that generates a new G1 dataset with simulated contact forces. Across both robots, state fusion outperforms the dedicated encoder. On G1, state fusion raises the success rate from 48\% with standard fine-tuning to 94\%, whereas the dedicated encoder reaches 74\%. On GR1, the dedicated encoder performs worse than standard fine-tuning, whereas state fusion improves on it. These results suggest that a policy pretrained without contact can learn to use contact signals effectively through its existing proprioceptive input. Our G1 dataset is available at \url{https://huggingface.co/jnsungp}.

\end{abstract}
\section{Introduction}

Robot foundation policies are trained on large-scale demonstration data and can be adapted to new robots and tasks through fine-tuning~\cite{rt1,rt2,octo,openvla,pi0,pi05,gr00tn1}. However, the demonstrations used for pretraining consist mostly of camera images, language instructions, and proprioceptive states~\cite{oxe,droid}, so these policies are typically pretrained without contact signals~\cite{fuse,justaddforce}. Contact signals are important for manipulation because they show when the hand grasps an object, whether the grasp holds, and when the object slips or is released. Without a contact signal, a policy relies on images to detect these events, which is difficult when the hand occludes the object, as often happens in the egocentric view of a humanoid robot. A contact signal provides evidence of these events that does not depend on the image, and prior work shows that tactile and force sensing improve manipulation policies on contact-rich tasks~\cite{hato,vitac3d,rdp,forcemimic,foar}.

However, collecting demonstrations with contact at scale remains challenging, since it requires equipping robots with additional tactile or force sensors~\cite{vlatouch,justaddforce}. Moreover, most existing large-scale real-world datasets were collected without such sensors, so their contact signals cannot be recovered after collection~\cite{oxe,droid}. A practical alternative is therefore to add contact when fine-tuning the policy on a small amount of data from the target robot, which can be collected with the sensors of that robot. In this setting, the pretrained policy has never observed contact, and a key challenge is to exploit the new signal without degrading the pretrained policy. This raises a question:
\begin{quote}
{\em Where should contact enter a policy that was pretrained without it?}
\end{quote}

Recent work has added tactile and force signals to pretrained foundation policies, either by fine-tuning them with new tactile encoders~\cite{fuse}, by feeding tactile signals to the vision-language backbone as tokens~\cite{tla,vtla,tactilevla,omnivtla,mla}, or by refining their actions with a separate tactile controller~\cite{vlatouch}. For force and torque, ForceVLA~\cite{forcevla} fuses force--torque measurements with vision-language features, and TA-VLA~\cite{tavla} shows that the performance of a policy depends on where joint torque enters it. These results suggest that how a new signal is added to a pretrained policy matters, not only whether it is used.

In this work, we extend this line of research to humanoid robots with multi-fingered hands and study how to add hand contact to GR00T N1.5~\cite{gr00tn15} during fine-tuning (Figure~\ref{fig:contact_methods}). We compare state fusion, which concatenates the contact signal to the proprioceptive state so that contact enters through the state encoder that the policy already uses, with a dedicated contact encoder, which gives the encoded contact signal to the action head as a new token. We evaluate both methods on two humanoid robots, the Fourier GR1 and the Unitree G1, and fine-tune a separate policy for each robot.

Evaluating both methods requires humanoid demonstrations that record contact, which existing datasets rarely provide~\cite{gr00tn1,gr00tteleopg1,unitreeg1,humanoideveryday}. For GR1, we augment the public simulation dataset~\cite{gr00tn1} with a binary contact signal computed from collisions between the hand and the object in the simulator. For G1, we build a simulation pipeline that plans pick-and-place motions with cuRobo~\cite{curobo} and records the contact forces at the fingertips and palms of both hands, and we release the resulting dataset of about 1K demonstrations.

We evaluate each policy with 100 closed-loop rollouts per task on three GR1 tasks and one G1 task. On G1, state fusion raises the success rate of standard fine-tuning from \textbf{48\%} to \textbf{94\%}, whereas the dedicated encoder reaches \textbf{74\%}. On GR1, where the contact signal is a single binary value, state fusion improves over standard fine-tuning on all three tasks, whereas the dedicated encoder falls below it on all three. Finally, we also fine-tune the policy with depth, another widely used modality. Contact through state fusion outperforms depth on G1, and on GR1, combining the two matches or exceeds either modality alone on all three tasks.

Together, these results show that a pretrained policy can learn to use contact through fine-tuning, and that for GR00T N1.5, fusing contact with the proprioceptive state that the policy already uses is more effective than encoding it with a dedicated encoder.

Our contributions are as follows:
\begin{itemize}
    \item We compare state fusion and a dedicated contact encoder as two ways of adding contact to a pretrained robot foundation policy during fine-tuning.
    \item We build a simulation pipeline that generates Unitree G1 demonstrations with fingertip and palm contact forces, and we release a dataset of about 1K demonstrations.
    \item We adapt GR00T N1.5 to GR1 and G1 with different contact signals and evaluate both methods on each robot.
\end{itemize}

\section{Related Work}

\subsection{Robot Foundation Policies}

Robot foundation policies are trained on demonstrations collected from many robots and tasks~\cite{oxe,droid}, and some also use demonstrations generated automatically in simulation~\cite{mimicgen,dexmimicgen,gr00tn1}. RT-1~\cite{rt1} and RT-2~\cite{rt2} predict discretized actions with a transformer and a vision-language model, respectively. Octo~\cite{octo} and OpenVLA~\cite{openvla} are open-source policies pretrained on Open X-Embodiment data. More recent policies generate continuous action chunks with diffusion~\cite{diffusionpolicy,rdt} or flow matching~\cite{flowmatching,pi0,pi05}. To cover robots with different sensors and action spaces, several policies place embodiment-specific input and output modules around a shared backbone~\cite{crossformer,hpt,gr00tn1}.

GR00T N1~\cite{gr00tn1} is a foundation policy for humanoid robots. A vision-language model encodes images and language, and a diffusion transformer (DiT)~\cite{dit} action head trained with flow matching generates actions. We use its successor, GR00T N1.5~\cite{gr00tn15}, as the base policy. In GR00T N1.5, the vision-language model is frozen, and the DiT cross-attends to its embeddings while processing the noised actions and the state, which is embedded by an embodiment-specific state encoder. These policies are pretrained with images, language instructions, and proprioceptive states as inputs, and contact signals are not among them. Our focus is on how to add contact to such a policy during fine-tuning.

\subsection{Tactile and Force Sensing for Manipulation Policies}

Tactile and force sensing have long been used for learning contact-rich manipulation. Earlier work learns multimodal representations that combine vision with tactile or force signals for policy learning~\cite{makingsense,m3l}. Recent imitation learning systems train visuo-tactile policies with multi-fingered hands or dense tactile sensors~\cite{hato,vitac3d}, and use tactile or force feedback for delicate grasping and for reactive or compliant control~\cite{justaddforce,rdp,forcemimic,acp,foar}. In these policies, the tactile or force input is used from the start of training. Pretrained tactile encoders, in turn, transfer across vision-based tactile sensors~\cite{sparsh,t3,anytouch}.

More recent work adds tactile or force signals to pretrained robot foundation policies, and Xie and Correll~\cite{forcefulsurvey} survey this direction. FuSe~\cite{fuse} fine-tunes generalist policies with tactile and audio encoders, using auxiliary losses that ground the new sensors in language. Tactile vision-language-action models feed tactile signals to the vision-language backbone as tokens~\cite{tla,vtla,tactilevla,omnivtla,mla}. VLA-Touch~\cite{vlatouch} does not fine-tune the base policy and instead refines its actions with a separate tactile controller. ForceVLA~\cite{forcevla} fuses six-axis force--torque measurements with vision-language features through a mixture-of-experts module. TA-VLA~\cite{tavla} studies where joint torque should enter $\pi_0$~\cite{pi0}, and compares giving it to the vision-language backbone as a token, placing it in the zero-padded dimensions of the state, and giving it to the action expert as a separate token. Our work follows this line of research and studies hand contact on humanoid robots with multi-fingered hands, where contact is a binary signal or a vector of fingertip and palm forces.

\subsection{Adding New Modalities to Pretrained Models}

A new input can disrupt what a pretrained model has learned, so methods that add a modality are usually designed to preserve the pretrained model. Flamingo~\cite{flamingo} connects frozen vision and language models through new cross-attention layers whose gates are initialized to zero. ControlNet~\cite{controlnet} adds spatial conditioning to a frozen text-to-image diffusion model through zero-initialized layers. For robot policies, a widely studied new modality is depth or 3D information. Depth Helps~\cite{depthhelps} uses depth when fine-tuning a pretrained RGB-based policy and predicts it from RGB at deployment. PointVLA~\cite{pointvla} freezes the pretrained action expert and injects point-cloud features into only a few of its blocks. Other models build depth or 3D inputs into the policy itself~\cite{vla3d,spatialvla,depthvla}. In this work, we add contact to a pretrained policy either through the state encoder that it already has or through a dedicated contact encoder. We also fine-tune the policy with depth, using depth maps estimated with ZoeDepth~\cite{zoedepth} for GR1.

\section{Method}
\label{sec:method}

\begin{figure*}[t]
    \centering
    \includegraphics[width=\textwidth]{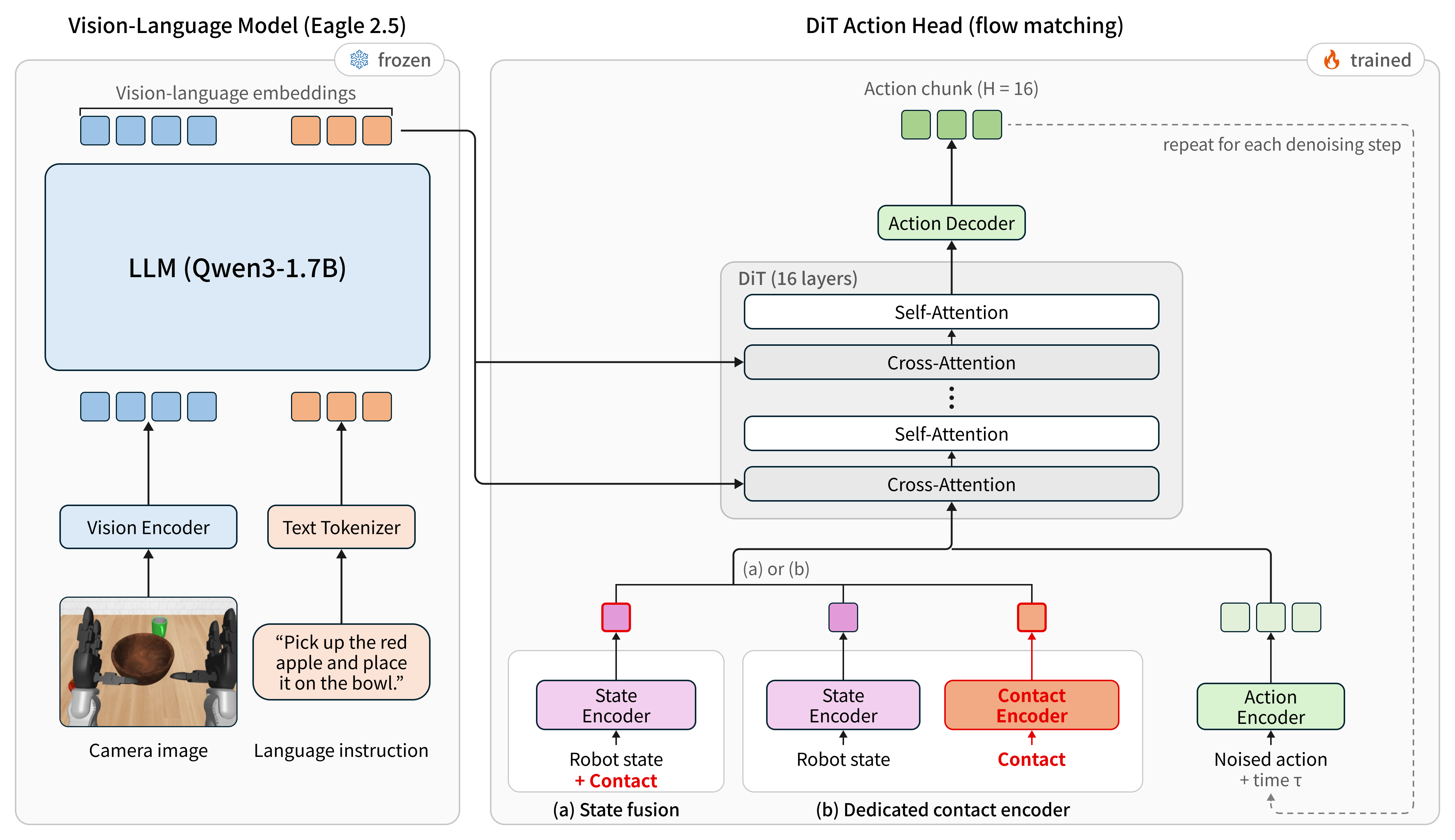}
    \caption{\textbf{Two ways of adding contact to GR00T N1.5.} The vision-language model is frozen, and the DiT action head is fine-tuned. (a) State fusion concatenates the contact signal to the robot state, so that contact enters through the existing state token. (b) A dedicated contact encoder maps the contact signal to a separate contact token, which is placed immediately after the state token. Each fine-tuned policy uses either (a) or (b).}
    \label{fig:contact_methods}
\end{figure*}

In this section, we first formulate the problem and describe the pretrained policy (Section~\ref{sec:problem}). We then present state fusion (Section~\ref{sec:state_fusion}) and the dedicated contact encoder (Section~\ref{sec:contact_encoder}), the two methods that we compare (Figure~\ref{fig:contact_methods}).

\subsection{Problem Formulation}
\label{sec:problem}

We consider fine-tuning a pretrained robot foundation policy on a dataset $\mathcal{D}$ of demonstrations from a target robot. At each timestep $t$, the policy receives camera images $o_t$, a language instruction $\ell$, and the proprioceptive state $s_t$, and it generates an action chunk $A_t = (a_t, \dots, a_{t+H-1})$ of $H = 16$ future actions. Standard fine-tuning learns $\pi_\theta(A_t \mid o_t, \ell, s_t)$. In our setting, the demonstrations also contain a contact signal $c_t$ that describes the contact between the hand and the manipulated object, and we learn $\pi_\theta(A_t \mid o_t, \ell, s_t, c_t)$.

We use GR00T N1.5~\cite{gr00tn15} as the pretrained policy. A frozen vision-language model encodes the images and the instruction into vision-language embeddings $\phi_t$, and a DiT~\cite{dit} action head generates the action chunk. The action head consists of the DiT and the embodiment-specific state encoder, action encoder, and action decoder. The state encoder maps the state $s_t$ to a state token $z^{s}_t$, and the action encoder maps a noised action chunk $A^{\tau}_t$ and its time $\tau$ to action tokens $z^{a}_t$. The action decoder maps the outputs of the DiT to the flow-matching velocity. The input sequence of the DiT is
\begin{equation}
X_t = [\, z^{s}_t,\; q,\; z^{a}_t \,],
\label{eq:dit_input}
\end{equation}
where $q$ denotes 32 learnable query tokens, which GR00T N1.5 uses for future latent representation alignment (FLARE)~\cite{flare} during pretraining. Within the DiT, self-attention over the tokens in $X_t$ alternates with cross-attention to the vision-language embeddings $\phi_t$.

The action head is trained with flow matching~\cite{flowmatching}. Given a demonstrated action chunk $A_t$, noise $\epsilon \sim \mathcal{N}(0, I)$, and a time $\tau \in [0, 1]$, the noised action chunk is $A^{\tau}_t = \tau A_t + (1 - \tau)\epsilon$, and the action head $V_\theta$ is trained to predict the velocity $A_t - \epsilon$ with the loss
\begin{equation}
\mathcal{L}(\theta) = \mathbb{E}_{\mathcal{D},\tau,\epsilon} \left\| V_\theta(\phi_t, A^{\tau}_t, \tau, z^{s}_t) - (A_t - \epsilon) \right\|^2 .
\label{eq:fm}
\end{equation}
At inference, the policy starts from Gaussian noise at $\tau = 0$ and integrates the predicted velocity with the Euler method, $A^{\tau+\Delta}_t = A^{\tau}_t + \Delta\, V_\theta(\phi_t, A^{\tau}_t, \tau, z^{s}_t)$, over four steps with $\Delta = 1/4$.

Because GR00T N1.5 is pretrained without contact, it has no input for $c_t$. The contact signal can therefore enter the policy either through an existing input or through a new one, and we study which of the two is more effective.

We consider two simulated humanoid robots with multi-fingered hands, the Fourier GR1 and the Unitree G1. The two robots differ in their contact signals, a binary value for GR1 and eight fingertip and palm contact forces for G1 (Section~\ref{sec:data}). Because they also differ in their state and action spaces, we fine-tune a separate policy for each robot from the same pretrained checkpoint.

\subsection{State Fusion}
\label{sec:state_fusion}

A simple way to add contact without changing the architecture of the pretrained policy is to treat it as part of the proprioceptive state, which the policy already encodes. In state fusion, we concatenate the contact signal to the state vector and pass the result through the existing state encoder (Figure~\ref{fig:contact_methods}a),
\begin{equation}
\tilde{s}_t = [\, s_t,\; c_t \,], \qquad z^{s}_t = \text{StateEnc}(\tilde{s}_t).
\label{eq:state_fusion}
\end{equation}
Since GR00T N1.5 pads the state with zeros to a fixed maximum dimension before the state encoder, the contact dimensions, one for GR1 and eight for G1, occupy inputs that are otherwise padded. State fusion therefore adds neither a new module nor a new token, and the input sequence in Equation~\eqref{eq:dit_input} keeps the structure used in pretraining, with contact changing only the content of the state token.

\subsection{Dedicated Contact Encoder}
\label{sec:contact_encoder}

In contrast to state fusion, the dedicated contact encoder treats contact as a new modality with its own encoder and token (Figure~\ref{fig:contact_methods}b). The state encoder receives the original state $s_t$, and a new contact encoder, a multilayer perceptron with the same architecture as the state encoder, maps the contact signal to a contact token,
\begin{equation}
z^{c}_t = \text{ContactEnc}(c_t).
\label{eq:contact_enc}
\end{equation}
We insert this token immediately after the state token in the input sequence of the DiT,
\begin{equation}
X_t = [\, z^{s}_t,\; z^{c}_t,\; q,\; z^{a}_t \,],
\label{eq:contact_tokens}
\end{equation}
while the vision-language embeddings $\phi_t$ remain the input to cross-attention. Through self-attention, each action token can then attend to the state token and the contact token separately. Because the contact encoder is a new module, it is trained from scratch during fine-tuning, and the DiT receives a token that it never observed during pretraining.

Both methods use the same contact signal and the same training objective in Equation~\eqref{eq:fm}, in which $V_\theta$ also takes the contact token $z^{c}_t$ as input when the dedicated contact encoder is used. They differ in where contact enters the policy, through the existing state token in state fusion and through a new token from a new encoder in the dedicated contact encoder.

\section{Experiments}
\label{sec:experiments}

We evaluate whether adding contact during fine-tuning improves a pretrained policy over standard fine-tuning, and which of the two methods is more effective. After describing the datasets and contact signals (Section~\ref{sec:data}) and the training and evaluation setup (Section~\ref{sec:exp_setup}), we compare the two methods on GR1 (Section~\ref{sec:exp_gr1}) and on G1 (Section~\ref{sec:exp_g1}). Finally, we also fine-tune the policy with depth, another widely used modality (Section~\ref{sec:exp_depth}).

\subsection{Datasets and Contact Signals}
\label{sec:data}

We evaluate on three GR1 tasks and one G1 task, all of which require the robot to grasp an object, hold it during transport, and release it at the target. For GR1, we use three tasks from the public simulation dataset released with GR00T N1~\cite{gr00tn1}, which are \textit{Pick-and-Place (PnP) Can to Drawer Close}, \textit{PnP Potato to Microwave Close}, and \textit{PnP Wine to Cabinet Close}, in which the robot places an object in a drawer, a microwave, or a cabinet and then closes it. Since this dataset provides RGB images, proprioceptive states, and actions but no contact signal, we label each timestep with a binary contact signal, setting $c_t = 1$ when the hand collides with the manipulated object in the simulator and $c_t = 0$ otherwise. This signal indicates whether contact occurs, but not where on the hand it occurs or how strong it is.

\begin{figure}[t]
    \centering
    \begin{tabular}{@{}c@{\hspace{0.03\linewidth}}c@{}}
        \includegraphics[width=0.485\linewidth]{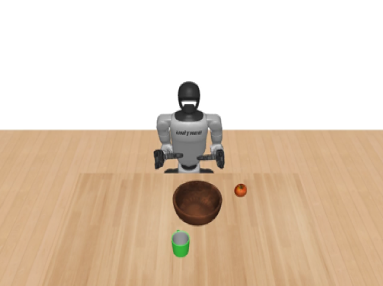} &
        \includegraphics[width=0.485\linewidth]{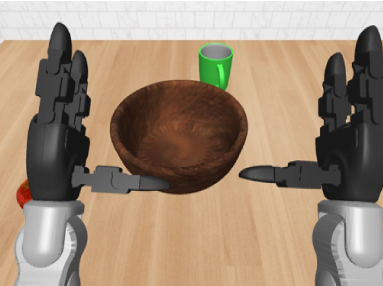} \\
        {\footnotesize (a)} & {\footnotesize (b)}
    \end{tabular}
    \caption{\textbf{Simulation scene for G1 data generation.} (a) Front view of the scene with the G1, the apple, and the bowl. (b) Egocentric view from the head camera of the G1 during a demonstration.}
    \label{fig:g1_dataset_views}
\end{figure}

For G1, we generate demonstrations for the task \textit{Pick Apple to Bowl} in simulation, because few public G1 datasets include contact. The datasets released by NVIDIA and Unitree~\cite{gr00tteleopg1,unitreeg1} do not record contact between the hand and the object, and Humanoid Everyday~\cite{humanoideveryday} is among the few that record tactile signals, collected on real robots. We build the scene in MuJoCo~\cite{mujoco} with assets from RoboCasa~\cite{robocasa}, where the G1 with two three-fingered hands stands in front of a table with an apple and a bowl (Figure~\ref{fig:g1_dataset_views}a), and the language instruction is ``Pick up the red apple and place it on the bowl.'' Instead of collecting demonstrations by teleoperation, we use cuRobo~\cite{curobo}, a GPU-accelerated motion planner, to solve inverse kinematics (IK) and plan collision-free joint trajectories that satisfy the grasping and placement constraints of the task. Executing these trajectories in simulation and keeping only the successful rollouts yields about 1K demonstrations. Each demonstration contains egocentric RGB images from the head camera, the upper-body joint positions as the state, the target joint positions as the action, and the contact forces of both hands.

\begin{table*}[t]
\centering
\caption{Contact Augmentation on GR1 and G1}
\label{tab:contact}
\begin{threeparttable}
\begin{tabular}{lcccccc}
\toprule
 & \multicolumn{4}{c}{GR1} & \multicolumn{2}{c}{G1} \\
\cmidrule(lr){2-5} \cmidrule(lr){6-7}
Method & Can & Potato & Wine & MSE $\downarrow$ & Apple to Bowl & MSE $\downarrow$ \\
\midrule
Zero-shot                     & 18.0 & 18.0 & 15.0 & 0.054825 & 0.0  & 0.35719 \\
Standard fine-tuning          & 51.0 & 49.0 & 46.0 & 0.021566 & 48.0 & 0.031716 \\
+ Contact (dedicated encoder) & 14.0 & 15.0 & 10.0 & 0.042569 & 74.0 & 0.024407 \\
+ Contact (state fusion)      & \textbf{56.0} & \textbf{51.0} & \textbf{51.0} & \textbf{0.019662} & \textbf{94.0} & \textbf{0.018623} \\
\bottomrule
\end{tabular}
\begin{tablenotes}[flushleft]\footnotesize\setlength{\labelsep}{0pt}
\item[] The Can, Potato, Wine, and Apple to Bowl columns report the success rate (\%) over 100 rollouts on \textit{PnP Can to Drawer Close}, \textit{PnP Potato to Microwave Close}, \textit{PnP Wine to Cabinet Close}, and \textit{Pick Apple to Bowl}. The GR1 MSE is measured on \textit{PnP Can to Drawer Close}.
\end{tablenotes}
\end{threeparttable}
\end{table*}

The G1 contact signal $c_t \in \mathbb{R}^{8}$ contains, for each hand, the contact force at the thumb, index, and middle fingertips and at the palm. We approximate contact force using the penetration depth reported by MuJoCo, with $f = k \max(0, -\delta)$, where $\delta$ is the signed contact distance and $k$ is a constant.

\begin{figure}[t]
    \centering
    \includegraphics[width=0.95\linewidth]{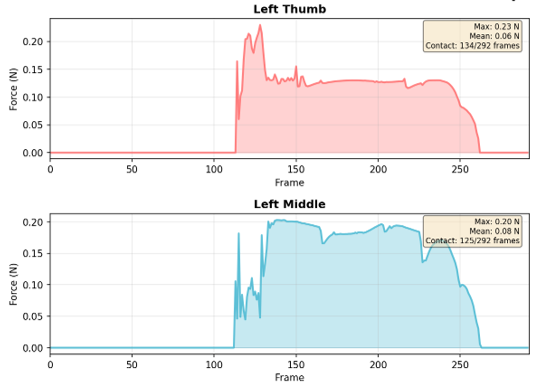}
    \caption{\textbf{Simulated contact forces in one G1 demonstration.} The forces at the left thumb and the left middle fingertip are zero during the approach, rise when the fingers close on the apple, and return to zero when the apple is released.}
    \label{fig:contact_force_plot}
\end{figure}

As shown in Figure~\ref{fig:contact_force_plot}, the contact forces at two fingertips are zero while the hand approaches the apple, rise when the fingers close on it, remain while it is transported, and return to zero when it is released. The contact signal thus indicates when contact with the object begins and ends, which is difficult to observe in the image when the hand occludes the object.

\subsection{Training and Evaluation}
\label{sec:exp_setup}

All policies start from the same pretrained GR00T-N1.5-3B checkpoint~\cite{gr00tn15}. We compare the pretrained policy without fine-tuning (\textit{Zero-shot}), the policy fine-tuned on the data of the target robot without contact (\textit{Standard fine-tuning}), and the policies fine-tuned with contact through state fusion and through the dedicated contact encoder (\textit{+ Contact (state fusion)} and \textit{+ Contact (dedicated encoder)}). On G1, all fine-tuned policies use the same demonstrations and differ only in their inputs and in how contact is added.

All fine-tuned policies are trained with the official GR00T fine-tuning framework for 20K steps with a batch size of 32, and we train five models with different random seeds for each policy. As in the official fine-tuning setup, the vision-language model, including its vision encoder, is frozen, and the action head is fine-tuned, including the embodiment-specific state and action encoders, the action decoder, and the DiT. The contact encoder, when used, is trained together with the rest of the action head, and for the RGB-D policies in Section~\ref{sec:exp_depth}, the expanded patch embedding of the vision encoder is also trained.

\begin{figure}[t]
    \centering
    \includegraphics[width=0.485\linewidth]{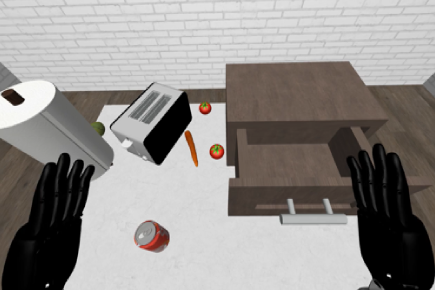}\hfill
    \includegraphics[width=0.485\linewidth]{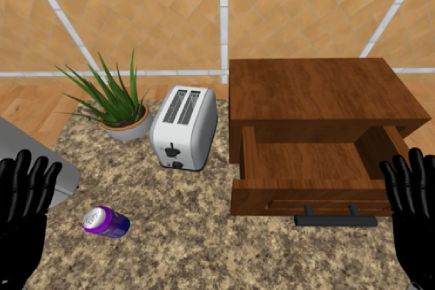}
    \caption{\textbf{Simulation environments for closed-loop evaluation.} The object placement, textures, lighting, and robot configuration are randomized in each rollout.}
    \label{fig:online_inf}
\end{figure}

We evaluate each policy with two metrics. Our main metric is the success rate in closed-loop rollouts, which we run in simulation environments built on robosuite~\cite{robosuite} and RoboCasa~\cite{robocasa} with randomized initial object positions and robot configurations (Figure~\ref{fig:online_inf}). For each task, we run 20 rollouts with each of the five fine-tuned models, which gives 100 rollouts in total, and we also run 100 rollouts with the zero-shot policy. We report the percentage of rollouts in which the task is completed.

Our second metric is the action prediction error on 100 held-out demonstrations. For each demonstration, the policy predicts an action chunk of $H = 16$ steps from the recorded observation at step 0, then predicts the next chunk from the recorded observation at step 16, and so on. We concatenate the predicted chunks, keep the first $T = 150$ steps, and compute the mean squared error (MSE) between the predicted actions $\hat{a}_t$ and the demonstrated actions $a_t$ over all joints,
\begin{equation}
\text{MSE} = \frac{1}{T J} \sum_{t=0}^{T-1} \left\| \hat{a}_t - a_t \right\|_2^2 ,
\label{eq:mse}
\end{equation}
where $J$ is the number of joints, which is 26 for GR1 and 28 for G1. We report the average over the 100 demonstrations and, for the fine-tuned policies, over the five models. For GR1, we report the MSE on \textit{PnP Can to Drawer Close}. Because each chunk is predicted from a recorded observation, the MSE measures how well the policy reproduces the demonstrations, but it does not capture errors that accumulate during closed-loop execution. We therefore use it as a complement to the success rate.

\begin{table*}[t]
\centering
\caption{Fine-Tuning with Depth}
\label{tab:depth}
\begin{threeparttable}
\begin{tabular}{lcccccc}
\toprule
 & \multicolumn{4}{c}{GR1} & \multicolumn{2}{c}{G1} \\
\cmidrule(lr){2-5} \cmidrule(lr){6-7}
Method & Can & Potato & Wine & MSE $\downarrow$ & Apple to Bowl & MSE $\downarrow$ \\
\midrule
Standard fine-tuning             & 51.0 & 49.0 & 46.0 & 0.021566 & 48.0 & 0.031716 \\
+ Contact (state fusion)         & 56.0 & \textbf{51.0} & 51.0 & 0.019662 & \textbf{94.0} & \textbf{0.018623} \\
+ Depth                          & 61.0 & 49.0 & 59.0 & 0.015234 & 82.0 & 0.022596 \\
+ Contact (state fusion) + Depth & \textbf{63.0} & \textbf{51.0} & \textbf{61.0} & \textbf{0.011547} & -- & -- \\
\bottomrule
\end{tabular}
\begin{tablenotes}[flushleft]\footnotesize\setlength{\labelsep}{0pt}
\item[] The columns are the same as in Table~\ref{tab:contact}. The combination of contact and depth is not evaluated on G1.
\end{tablenotes}
\end{threeparttable}
\end{table*}

\subsection{Results on GR1}
\label{sec:exp_gr1}

On GR1, standard fine-tuning substantially improves over the zero-shot policy on all three tasks (Table~\ref{tab:contact}). Adding contact through state fusion improves the policy further on every task and also lowers the action prediction error, although the gains are modest. In contrast, the dedicated contact encoder degrades the policy. It falls below standard fine-tuning, and also below the zero-shot policy, on all three tasks, and its action prediction error is higher than that of standard fine-tuning. The same contact signal thus improves the policy when it is fused with the state, but degrades it when it enters the DiT as a new token.

\subsection{Results on G1}
\label{sec:exp_g1}

On G1, where the contact signal consists of eight fingertip and palm forces, the zero-shot policy fails in every rollout, and standard fine-tuning on about 1K demonstrations succeeds in only about half of them (Table~\ref{tab:contact}). Unlike on GR1, both methods improve over standard fine-tuning, and state fusion again outperforms the dedicated encoder, succeeding in nearly all rollouts. The action prediction error follows the same order. Across both robots, state fusion is therefore the more effective way to add contact.

Although the gain from contact is larger on G1 than on GR1, the two settings differ not only in the contact signal, eight forces instead of a single bit, but also in the robot, the task, and the data, so we cannot attribute this difference to the contact signal alone.

\subsection{Fine-Tuning with Depth}
\label{sec:exp_depth}

\begin{figure}[t]
    \centering
    \begin{tabular}{@{}c@{\hspace{0.03\linewidth}}c@{}}
        \includegraphics[width=0.485\linewidth]{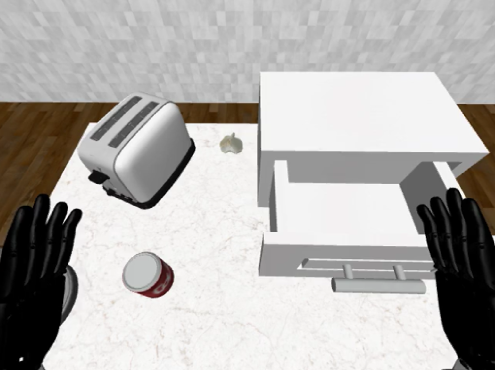} &
        \includegraphics[width=0.485\linewidth]{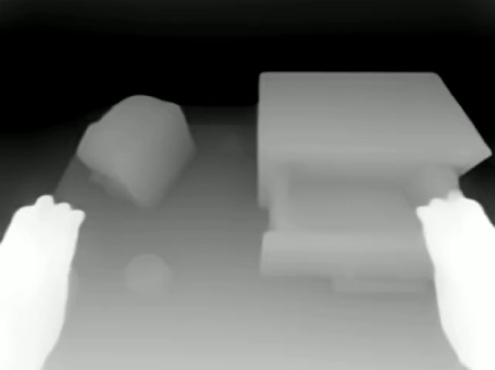} \\
        {\footnotesize (a)} & {\footnotesize (b)}
    \end{tabular}
    \caption{\textbf{RGB image and estimated depth for a GR1 demonstration.} (a) RGB image from the GR1 dataset. (b) Metric depth map of (a) estimated by ZoeDepth.}
    \label{fig:rgb_depth_example}
\end{figure}

Finally, we also fine-tune the policy with depth, another modality that is widely used in robot policies~\cite{depthhelps,pointvla,spatialvla,depthvla}. We add depth as an image, so it enters the policy through the vision encoder rather than through the action head as contact does. Since the public GR1 dataset contains only RGB images, we estimate a metric depth map for each frame with ZoeDepth~\cite{zoedepth}, a monocular depth estimator (Figure~\ref{fig:rgb_depth_example}b), whereas for G1 the simulator renders the depth image from the head camera. The vision encoder of the Eagle 2.5 vision-language model in GR00T N1.5~\cite{gr00tn15} embeds each $p \times p$ RGB patch with a linear projection. To take depth as an input, we concatenate the depth map to the RGB channels and expand this projection from three to four channels, so that each patch $P_t$ of the four-channel RGB-D image $I_t$ is embedded as
\begin{equation}
e_t = W_{\text{patch}}\, \text{Flatten}(P_t) + b_{\text{patch}},
\label{eq:patch}
\end{equation}
where $W_{\text{patch}} \in \mathbb{R}^{d \times (p \cdot p \cdot 4)}$ and $d$ is the embedding dimension. We initialize the weights of the depth channel to the mean of the pretrained weights of the three RGB channels,
\begin{equation}
W_{\text{patch}}^{(D)} = \frac{1}{3}\left( W_{\text{patch}}^{(R)} + W_{\text{patch}}^{(G)} + W_{\text{patch}}^{(B)} \right).
\label{eq:depth_init}
\end{equation}
We train the expanded patch embedding, while the rest of the vision encoder and the language model remain frozen.

Table~\ref{tab:depth} shows that depth and contact are complementary on GR1. Depth improves standard fine-tuning more than contact on the Can and Wine tasks and lowers the action prediction error further, but it does not improve the success rate on the Potato task, whereas contact does. Adding contact through state fusion on top of depth matches or exceeds the best single modality on all three tasks and gives the lowest action prediction error, which suggests that contact provides information that depth does not. On G1, depth outperforms the dedicated contact encoder but not state fusion, and the action prediction error follows the same order.


\section{Conclusion}
\label{sec:conclusion}

We studied how to add contact to a robot foundation policy that was pretrained without it. We compared state fusion, which concatenates the contact signal to the proprioceptive state, with a dedicated contact encoder, which gives the encoded contact signal to the action head of GR00T N1.5 as a new token, and we released a Unitree G1 dataset with simulated fingertip and palm contact forces. On both GR1 and G1, state fusion improved over standard fine-tuning and outperformed the dedicated encoder, which degraded the policy on GR1. More broadly, these results suggest that fusing contact with an input that the policy already uses can be a simple and effective way to add contact to a pretrained policy during fine-tuning.

Our study has limitations that suggest directions for future work. All experiments are conducted in simulation, where the binary GR1 signal and the penetration-based G1 forces differ from the signals of real tactile sensors. Evaluating both methods with real tactile sensors would address this limitation, and experiments on more tasks and with other foundation policies would also show how general our findings are.

\bibliographystyle{IEEEtran}
\bibliography{references}

\end{document}